\documentclass{article} 
\usepackage{iclr2018_conference,times}
\usepackage{hyperref}
\usepackage{url}

\usepackage{algorithm}
\usepackage{algorithmic}

\usepackage{bbm}
\usepackage{amsthm}
\usepackage{amsmath}
\usepackage{amssymb}
\usepackage{color}
\usepackage{xspace}
\usepackage{pgfplots, pgfplotstable}

\newcommand{\argmax}[1]{\underset{#1}{\operatorname{argmax}}}

\def\deriv{\mathrm{d}}

\renewcommand{\mid}[0]{\:\vert\:}

\def\tpcla{Trust-PCL (on-policy)\xspace}
\def\tpclb{Trust-PCL (off-policy)\xspace}

\def\expected{\mathbb{E}}
\def\ppi{\tilde{\pi}}
\def\pitheta{\pi_\theta}
\def\ppitheta{\pi_{\tilde\theta}}

\def\vphi{V_\phi}

\def\pistar{\pi^*}

\def\pistar{\pi^*}

\def\vstar{V^*}

\newcommand{\R}[1]{R}
\newcommand{\G}[1]{G}
\newcommand{\HH}{\mathbb{H}}
\newcommand{\EE}{\mathbb{G}}

\newcommand{\Ad}[1]{A}

\newcommand{\CC}[1]{C}

\def\calL{\mathcal{L}}

\def\calA{\mathcal{A}}

\def\eg{{\em e.g.,}}
\def\ie{{\em i.e.,}}

\newcommand{\figref}[1]{Figure~\ref{#1}}

\newcommand{\tabref}[1]{Table~\ref{#1}}

\def\temp{\tau}

\def\mro{O_{\text{ER}}}
\def\tmro{\overset{\sim}{O}_{\text{ER}}}

\def\ento{O_{\text{ENT}}}
\def\rento{O_{\text{RELENT}}}

\newcommand{\trans}[1]{{#1}^{\ensuremath{\mathsf{T}}}}
\newcommand{\kl}[2]{{\mathrm{KL}}\left(#1~\Vert~#2\right)}
\newcommand{\kld}[2]{D_{\mathrm{KL}}(#1,\,#2)}
\newcommand{\bg}[2]{D_{\mathrm{F}}(#1,\,#2)}

\newcommand{\comment}[1]{}
\newcommand{\opensource}{\url{https://github.com/tensorflow/models/tree/master/research/pcl_rl}}

\title{Trust-PCL: An Off-Policy \\Trust Region Method for Continuous Control}

\author{
Ofir Nachum, Mohammad Norouzi, Kelvin Xu, \& 
Dale Schuurmans\thanks{Also at the Department of Computing Science, University of Alberta, {\tt daes@ualberta.ca}}
\\
\texttt{\{ofirnachum,mnorouzi,kelvinxx,schuurmans\}@google.com} \\
Google Brain
}

\iclrfinalcopy 

\begin{document}

\maketitle

\begin{abstract}

Trust region methods, such as TRPO, are often used to stabilize policy
optimization algorithms in reinforcement learning (RL). While current
trust region strategies are effective for continuous control, they
typically require a large amount of on-policy interaction with 
the environment.  To address this problem, we propose an
off-policy trust region method, \emph{Trust-PCL},
which
exploits an observation that
the optimal policy and state values of a
maximum reward objective with a relative-entropy
regularizer satisfy a set of multi-step pathwise
consistencies along any path.
The introduction of
relative entropy regularization allows
Trust-PCL
to maintain
optimization stability while exploiting off-policy data to improve
sample efficiency. When evaluated on a number of continuous control
tasks, Trust-PCL significantly 
improves the solution quality and sample efficiency
of TRPO.\footnote{An implementation of Trust-PCL is available at \opensource}

\comment{
Trust region methods such as TRPO are a common way to stabilize
the training of reinforcement learning (RL) agents.
While these methods add a necessary stability to policy-based
algorithms, they require a large number of experience collected
from the environment, often leading to a sample efficiency
that is infeasible for real-world systems.
Towards the aim of maintaining stability while improving
sample efficiency, we derive an off-policy trust region
algorithm.  
The algorithm is the result of observing that
the optimal policy and state values of a
maximum reward objective with a relative-entropy
regularizer satisfy a set of multi-step pathwise
consistencies along any path.
We evaluate our algorithm on a number of standard control tasks.
We find that our algorithm can solve these tasks
and, moreover, that it can improve upon TRPO
in terms of both average reward and sample efficiency.
}

\end{abstract}

\section{Introduction}

The goal of model-free reinforcement learning (RL) is to optimize an
agent's behavior policy through trial and error interaction with a
black box environment.  Value-based RL algorithms such as
Q-learning~\citep{qlearning} and policy-based algorithms such as
actor-critic~\citep{konda2000actor} have achieved well-known successes
in environments with enumerable action spaces and predictable but
possibly complex dynamics, \eg\ as in Atari games~\citep{atari, dqn,
  mnih2016asynchronous}. However, when applied to environments with more
sophisticated action spaces and dynamics (\eg~continuous
control and robotics), success has been far more limited.

In an attempt to improve the applicability of Q-learning to continuous
control, \citet{ddpg1} and \citet{ddpg2} developed an off-policy algorithm
DDPG, leading to promising results on continuous control
environments. That said, current off-policy methods including DDPG
often improve data efficiency at the cost of optimization stability.
The behaviour of DDPG is known to be highly
dependent on hyperparameter selection and
initialization~\citep{lmetz}; even when using optimal hyperparameters,
individual training runs can display highly varying outcomes.

On the other hand, in an attempt to improve the stability and convergence
speed of policy-based RL methods, \citet{kakade2002natural} developed
a natural policy gradient algorithm based on~\citet{amari1998natural},
which subsequently led to the development of trust region
policy optimization (TRPO)~\citep{trpo}.
TRPO has shown strong empirical performance on difficult continuous
control tasks often outperforming value-based methods like DDPG.
However, a major drawback is that such methods are not able to exploit
off-policy data and thus require a large amount of on-policy
interaction with the environment, making them impractical
for solving challenging real-world problems.  \comment{However, these methods
  require an immense amount of on-policy interaction with the
  environment, usually resulting in sample efficiencies that are
  impractical for solving challenging real-world problems.  }

Efforts at combining the stability of trust region policy-based methods 
with the sample efficiency of value-based methods 
have focused on using off-policy data to better train a value estimate,
which can be used as a control variate for variance reduction
~\citep{qprop, gu2017interpolated}.
 
In this paper, we investigate an alternative approach 
to improving the sample efficiency of trust region policy-based RL methods.  
We exploit the key fact that, under entropy regularization,
the optimal policy and value function satisfy a set of pathwise
consistency properties along \emph{any} sampled path \citep{pcl},
which allows both on and off-policy data to be incorporated in an 
actor-critic algorithm, PCL.
The original PCL algorithm optimized an entropy regularized 
maximum reward objective and was evaluated on relatively
simple tasks.
Here we extend the ideas of PCL to achieve strong results
on standard, challenging continuous control benchmarks.
The main observation is that by 
alternatively augmenting the maximum reward objective
with a relative entropy regularizer,
the optimal policy and values still satisfy a certain set of pathwise 
consistencies along any sampled trajectory.
The resulting objective is equivalent to maximizing expected reward subject 
to a penalty-based constraint on divergence from a reference (\ie\ previous)
policy.
 
We exploit this observation to propose a new off-policy trust region 
algorithm, \emph{Trust-PCL}, that is able to exploit off-policy data to train 
policy and value estimates.
Moreover, we present a simple method for determining 
the coefficient on the relative entropy regularizer 
to remain agnostic to reward scale, 
hence ameliorating the task of hyperparameter tuning.  
We find that the incorporation of a relative entropy regularizer is
crucial for good and stable performance.
We evaluate Trust-PCL against TRPO,
and observe that Trust-PCL is able to solve difficult continuous control 
tasks, 
while improving the performance of TRPO both in terms of the final
reward achieved as well as sample-efficiency.  

\section{Related Work}
\comment{
Our work combines the stability of trust region methods
with the effectiveness of multi-step pathwise consistencies.
We highlight the relevant literature which introduced 
and previously utilized these two ideas.
}

{\bf Trust Region Methods.} 
\comment{Policy gradient and actor-critic
methods~\citep{Williams92,sutton2000policy,konda2000actor} 
are popular model-free RL algorithms for optimizing a parametric policy
using stochastic gradient descent. 
The gradient of a policy's expected reward
is estimated by drawing Monte Carlo samples from the environment,
which typically yields gradient estimates with extremely high variance. 
In practice, a small learning rate is required to ensure stable optimization.
}
Gradient descent is the predominant optimization method for 
neural networks. 
A gradient descent step is equivalent to solving a
trust region constrained optimization,
\begin{equation}
\mathrm{minimize}~\ell(\theta + \deriv \theta) \approx \ell(\theta) +
\nabla \trans{\ell(\theta)}{\deriv
  \theta}~~~~~~\text{s.\:t.}~~~~~~
\trans{\deriv \theta}{\deriv \theta}\le\epsilon~,
\label{eq:euc-search}
\end{equation}
which yields the locally optimal update $\deriv\theta = -\eta\nabla\ell(\theta)$
such that $\eta=\sqrt{\epsilon}/\|\nabla\ell(\theta)\|$;
hence by considering a Euclidean ball,
gradient descent assumes the parameters lie in a Euclidean space.

However, in machine learning,
particularly in the context of multi-layer neural network training, 
Euclidean geometry is not necessarily the best way to characterize 
proximity in parameter space.
It is often more effective to define an appropriate Riemannian metric
that respects the loss surface \citep{Amari2012book},
which allows much steeper descent directions to be identified within a local
neighborhood (\eg~\cite{amari1998natural,kfac}). 
Whenever the loss is defined in terms of a Bregman divergence between an 
(unknown) optimal parameter $\theta^*$ and model parameter $\theta$, 
\ie~$\ell(\theta) \equiv \bg{\theta^*}{\theta}$, 
it is natural to use the same divergence to form the trust region:
\begin{equation}
\mathrm{minimize}~\bg{\theta^*}{\theta + \deriv \theta}
~~~~~~\text{s.\:t.}~~~~~~\bg{\theta}{\theta + \deriv \theta} \le \epsilon~.
\label{eq:bg-search}
\end{equation}
\comment{
Such a constraint will be particularly effective in the vicinity of $\theta^*$,
since any update $\deriv \theta$ cannot increase the loss by more than
$\epsilon$ given that $\bg{\theta^*}{\theta + \deriv \theta} \le
\epsilon$ when $\theta\approx\theta^*$.
}

The natural gradient~\citep{amari1998natural} is a generalization of
gradient descent where the Fisher information matrix $F(\theta)$
is used to define the local geometry of the parameter space around $\theta$.
If a parameter update is constrained by
$\trans{\deriv \theta}F(\theta){\deriv \theta} \le \epsilon$, 
a descent direction of 
$\deriv \theta \equiv -\eta F(\theta)^{-1} \nabla \ell(\theta)$
is obtained.
This geometry is especially effective for optimizing the log-likelihood
of a conditional probabilistic model, where the objective is in fact
the KL divergence $\kld{\theta^*}{\theta}$.
The local optimization is,
\begin{equation}
\mathrm{minimize}~\kld{\theta^*}{\theta + \deriv \theta}
~~~~~~\text{s.\:t.}~~~~~~\kld{\theta}{\theta + \deriv \theta} \approx \trans{\deriv \theta}F(\theta){\deriv
  \theta}\le \epsilon~.
\label{eq:kl-search}
\end{equation}
Thus, natural gradient approximates the trust region by
$\kld{a}{b}\approx\trans{(a-b)}F(a)(a-b)$,
which is accurate up to a second order Taylor approximation. 
Previous
work~\citep{kakade2002natural,bagnell2003covariant,peters2008schaal,trpo}
has applied natural gradient to policy optimization,
locally improving expected reward subject to variants of 
$\trans{\deriv \theta}F(\theta){\deriv \theta}\le \epsilon$.  
Recently, TRPO~\citep{trpo,trpo_gae} has achieved
state-of-the-art results in continuous control by adding
several approximations to the natural gradient 
to make nonlinear policy optimization feasible.

Another approach to trust region optimization is given by 
proximal gradient
methods~\citep{proximal}.  
The class of proximal gradient methods most similar to 
our work are those that replace the hard constraint 
in~\eqref{eq:bg-search} with a penalty added to the objective.
These techniques have recently become popular in 
RL~\citep{acer,heess2017emergence,ppo}, although in terms of final reward performance
on continuous control benchmarks, 
TRPO is still considered 
to be the state-of-the-art.

\cite{rml2016} make the observation that \emph{entropy regularized} expected
reward may be expressed as a reversed KL divergence
$\kld{\theta}{\theta^*}$,
which suggests that an alternative to the constraint in~\eqref{eq:kl-search} should be used
when such regularization is present:
\begin{equation}
\mathrm{minimize}~\kld{\theta + \deriv \theta}{\theta^*}
~~~~~~\text{s.\:t.}~~~~~~\kld{\theta + \deriv \theta}{\theta} \approx
\trans{\deriv \theta}F(\theta+\deriv\theta){\deriv \theta}\le
\epsilon~.
\label{eq:rkl-search}
\end{equation}
Unfortunately, this update requires computing the Fisher matrix at the
endpoint of the update.
The use of $F(\theta)$ in previous work can be considered to be an approximation
when entropy regularization is present, 
but it is not ideal, particularly if $\deriv\theta$ is large. 
In this paper, by contrast, we demonstrate that the optimal $\deriv \theta$ 
under the reverse KL constraint 
$\kld{\theta + \deriv \theta}{\theta} \le \epsilon$
can indeed be characterized.
Defining the constraint in this way appears to be more
natural and effective than that of TRPO.

{\bf Softmax Consistency.~} To comply with the information geometry over
policy parameters, previous work has used the relative entropy 
(\ie~KL divergence) to regularize policy optimization;
resulting in a softmax relationship between the optimal policy and state 
values~\citep{reps, azar, azaretal11, fox, rawlik}
under single-step rollouts.  Our work is unique in that we leverage
consistencies over multi-step rollouts.

The existence of multi-step softmax consistencies has been noted by
prior work---first by~\citet{pcl} in the presence of entropy regularization.
The existence of the same consistencies with relative entropy
has been noted by~\citet{schulman2017equivalence}.
Our work presents multi-step consistency relations
for a hybrid relative entropy plus entropy regularized expected reward objective,
interpreting relative entropy regularization as a trust region constraint.  
This work is also distinct from prior work
in that the coefficient of relative entropy can be
automatically determined,
which we have found to be especially crucial in cases where the
reward distribution changes dramatically during training.

\comment{
Our work generalizes the recently proposed PCL algorithm~\citep{pcl}
by incorporating a relative entropy trust region.  PCL is an
off-policy actor-critic-like algorithm that simultaneously learns a
parameterized policy and a state value model by minimizing multi-step
softmax temporal consistency errors. The objective of PCL is entropy
regularized expected. We extend this objective by incorporating a
relative-entropy trust region that proves to be extremely effective.
Accordingly, we present Trust-PCL, which can be thought as an
off-policy trust region policy optimization method, which makes use of
multi-step bootstrapping.
}

Most previous work on softmax consistency (\eg~\cite{fox, azar, pcl})
have only been evaluated on relatively simple tasks,
including grid-world and discrete algorithmic environments. 
\citet{rawlik} conducted evaluations on simple variants
of the CartPole and Pendulum continuous control tasks.  
More recently,~\citet{haarnoja2017reinforcement} showed that 
soft Q-learning (a single-step special case of PCL) can succeed on more
challenging environments, such as a variant of the Swimmer task we 
consider below.
By contrast, this paper presents a successful application of the softmax 
consistency concept to difficult and standard continuous-control benchmarks, 
resulting in performance that is competitive with and in some cases
beats the state-of-the-art. 

\section{Notation \& Background}
\label{prelim}

We model an agent's behavior by a policy distribution
$\pi(a \mid s)$ over a set of
actions (possibly discrete or continuous).  
At iteration $t$, the agent encounters a state $s_t$ and
performs an action $a_t$ sampled from $\pi(a \mid s_t)$.  The
environment then returns a scalar reward $r_t\sim r(s_t, a_t)$ 
and transitions to the next state $s_{t+1}\sim \rho(s_t, a_t)$.  
When formulating expectations over actions, rewards, and state transitions
we will often omit the sampling distributions, $\pi$, $r$, and $\rho$,
respectively.

{\bf Maximizing Expected Reward.~}
The standard objective in RL is to maximize expected future
discounted reward.
We formulate this objective on a per-state basis
recursively as
\begin{equation}
\mro(s, \pi) = \expected_{a, r, s^\prime}\left[r + \gamma \mro(s^\prime, \pi)\right].
\label{eq:objmr}
\end{equation}
The overall, state-agnostic objective is the expected per-state
objective when states are sampled from interactions with the
environment:
\begin{equation}
\mro(\pi) = \expected_s[\mro(s, \pi)].
\end{equation}
Most policy-based algorithms, including
REINFORCE~\citep{williams1991function} and
actor-critic~\citep{konda2000actor}, aim to optimize $\mro$ given a
parameterized policy.

{\bf Path Consistency Learning (PCL).~} 
Inspired by
\citet{williams1991function}, \citet{pcl} augment the
objective $\mro$ in~\eqref{eq:objmr} with a discounted entropy
regularizer to derive an objective,
\begin{equation}
\ento(s, \pi) ~=~ \mro(s, \pi) + \temp \HH(s, \pi)~,
\label{eq:objent1}
\end{equation}
where $\tau \ge 0$ is a user-specified temperature parameter that
controls the degree of entropy regularization,
and the discounted entropy $\HH(s, \pi)$ is recursively defined as
\begin{equation}
\HH(s, \pi) ~=~ \expected_{a, s'} [-\log \pi(a\mid s) + \gamma \HH(s', \pi)]~.
\label{eq:entropy}
\end{equation}
Note that the objective $\ento(s, \pi)$ can then be re-expressed recursively as,
\begin{equation}
\ento(s, \pi) ~=~ \expected_{a,r,s'} [ r -\temp \log \pi(a\mid s) + \gamma \ento(s', \pi)]~.
\label{eq:objent}
\end{equation}

\citet{pcl} show that the optimal policy $\pistar$ for
$\ento$ and $\vstar(s) = \ento(s, \pistar)$ 
mutually satisfy a softmax temporal consistency constraint along any sequence 
of states $s_0,\dots,s_{d}$
starting at $s_0$ and a corresponding sequence of actions
$a_0,\dots,a_{d-1}$:
\begin{equation}
\vstar(s_0) = \underset{r_i, s_i}{\expected}
\left[\gamma^d \vstar(s_d) + \sum_{i=0}^{d-1} \gamma^i (r_i - \tau \log \pistar(a_i | s_i)) \right].
\label{eq:multi}
\end{equation}
This observation led to the development of the PCL algorithm, which
attempts to minimize squared error between the LHS and RHS of
\eqref{eq:multi} to simultaneously optimize parameterized $\pitheta$
and $\vphi$. Importantly, PCL is applicable to both on-policy and
off-policy trajectories.

{\bf Trust Region Policy Optimization (TRPO).~}
As noted, standard policy-based algorithms for maximizing $\mro$
can be unstable and require small learning rates for training.
To alleviate this issue, \citet{trpo} proposed to perform an iterative
trust region optimization to maximize $\mro$.
At each step, a prior policy $\ppi$ is used to 
sample a large batch of trajectories, then $\pi$ is subsequently
optimized to maximize $\mro$ while remaining within a constraint
defined by the average per-state KL-divergence with $\ppi$.
That is, at each iteration TRPO solves the constrained optimization
problem,
\begin{equation}
\underset{\pi}{\mathrm{maximize}}~\mro(\pi)~~~~~~\text{s.\:t.}~~~~~~\expected_{s \sim \ppi,\rho} [\,\kl{\ppi(-|s)}{\pi(-|s)}\,] \le \epsilon.
\end{equation}
\comment{
\begin{align}
& \max_{\pi} ~\mro(\pi) \\
\text{s.t.} ~&~ \expected_s [\,\kl{\ppi(a|s)}{\pi(a|s)}\,] \le \epsilon.
\label{eq:trpo}
\end{align}}
The prior policy is then replaced with the new policy $\pi$,
and the process is repeated.

\section{Method}

To enable more stable training and better exploit the natural 
information geometry of the parameter space, 
we propose to augment the entropy regularized
expected reward objective $\ento$ in~\eqref{eq:objent1} with a
discounted relative entropy trust region around a prior policy $\ppi$,
\begin{equation}
\underset{\pi}{\mathrm{maximize}}~\expected_s [\ento(\pi)]~~~~~~\text{s.\:t.}~~~~~~\expected_s [\EE(s, \pi, \ppi)] \le \epsilon~,
\label{eq:tpcl-constraint}
\end{equation}
where the discounted relative entropy is recursively defined as
\begin{equation}
\EE(s, \pi, \ppi) = 
\expected_{a, s^\prime} \left[\log \pi(a|s) - \log \ppi(a|s) + \gamma \EE(s^\prime, \pi, \ppi)\right].
\label{eq:kl-con}
\end{equation}
This objective attempts to maximize entropy regularized expected
reward while maintaining natural proximity to the previous policy.
Although previous work has separately proposed to use relative entropy and
entropy regularization,
we find that the two components serve different purposes, 
each of which is beneficial:
entropy regularization helps improve exploration, 
while the relative entropy improves stability
and allows for a faster learning rate.
This combination is a key novelty.

Using the method of Lagrange multipliers, we cast the constrained
optimization problem in \eqref{eq:kl-con} into maximization of the
following objective,
\begin{equation}
\rento(s, \pi) = \ento(s, \pi) - \lambda \EE(s, \pi, \ppi)~.
\label{eq:objrent}
\end{equation}
Again, the environment-wide objective is the expected per-state
objective when states are sampled from interactions with the
environment,
\begin{equation}
\rento(\pi) = \expected_s[\rento(s, \pi)].
\end{equation}

\comment{
We note that the trust region constraint used in our paper differs
from natural policy gradient and TRPO methods in two important
ways. First, the direction of KL is different as discussed in the
related work. Second, natural gradient methods define the constraint
as an average per-state local KL divergence. By contrast, the
constraint induced by our objective measures the divergence over whole
trajectories.  When the discounting factor in the relative entropy
$\gamma=1$, then the constraint reduces to the expected
trajectory-wise KL-divergence starting from states sampled from
interactions with the environment.
}

\subsection{Path Consistency with Relative Entropy}
A key technical observation is that the $\rento$ objective has
a similar decomposition structure to $\ento$, 
and one can cast $\rento$ as an entropy regularized expected reward objective 
with a set of transformed rewards, \ie~
\begin{equation}
\rento(s, \pi) = \tmro(s, \pi) + (\tau + \lambda)\HH(s, \pi),
\end{equation}
where $\tmro(s, \pi)$ is an expected reward objective on a transformed 
reward distribution function $\tilde{r}(s, a) = r(s, a) + \lambda\log\ppi(a | s)$.
Thus, in what follows, we derive a corresponding form of the
multi-step path consistency in~\eqref{eq:multi}.

Let $\pistar$ denote the optimal policy, defined as
$\pistar = \argmax{} \text{}_\pi ~\rento(\pi)$.
As in PCL~\citep{pcl}, this optimal policy may be expressed as
\begin{equation}
\pistar(a_t|s_t) = \exp\left\{ 
\frac{\expected_{\tilde{r}_t\sim\tilde{r}(s_t,a_t),s_{t+1}}[\tilde{r}_t + \gamma V^*(s_{t+1})] - V^*(s_t)}
{\tau + \lambda} 
\right\},
\label{eq:piopt}
\end{equation}
where $V^*$ are the softmax state values defined recursively as
\begin{equation}
V^*(s_t) = (\tau+\lambda) \log\int_{\calA} \exp\left\{
\frac{\expected_{\tilde{r}_t\sim\tilde{r}(s_t,a),s_{t+1}}[\tilde{r}_t + \gamma V^*(s_{t+1})]}
{\tau + \lambda}
\right\} \deriv a.
\end{equation}
We may re-arrange~\eqref{eq:piopt} to yield
\begin{eqnarray}
V^*(s_t) &=& \expected_{\tilde{r}_t\sim\tilde{r}(s_t,a_t),s_{t+1}}[\tilde{r}_t - (\tau + \lambda)\log\pistar(a_t|s_t) + \gamma V^*(s_{t+1})] \\
         &=& \expected_{r_t,s_{t+1}}[r_t - (\tau + \lambda)\log\pistar(a_t|s_t) + \lambda \log {\ppi(a_{t+i} | s_{t+i})} + \gamma V^*(s_{t+1})].
\end{eqnarray}
This is a single-step temporal consistency which may be 
extended to multiple steps
by further expanding $V^*(s_{t+1})$ on the RHS using the same identity.
Thus, in general we have
the following softmax temporal consistency constraint along any sequence of states defined 
by a starting state $s_t$ and a sequence of actions $a_{t},\ldots,a_{t+d-1}$:
\begin{equation}
\vstar(s_t) = \underset{r_{t+i}, s_{t+i}}{\expected}
\left[\gamma^d \vstar(s_{t+d}) + 
\sum_{i=0}^{d-1} \gamma^i \left(r_{t+i} - (\tau+\lambda) \log \pistar(a_{t+i} | s_{t+i}) + \lambda \log {\ppi(a_{t+i} | s_{t+i})} \right)\right].
\label{eq:multi2}
\end{equation}

\comment{
Let $\vstar(s)$ denote the optimal value of a state $s$, which is given by
$\vstar(s) = \max_{\pi} \rento(s, \pi)$,
and let $\pistar$ denote the optimal policy, defined as
$\pistar = \argmax{} \text{}_\pi ~\rento(\pi)$.
As in PCL, this optimal policy and its optimal value function satisfy a 
softmax temporal consistency constraint along any sequence of states defined 
by a starting state $s_0$ and a sequence of actions $a_{0},\ldots,a_{d-1}$:
\begin{equation}
\vstar(s_0) = \underset{r_i, s_i}{\expected}
\left[\gamma^d \vstar(s_d) + 
\sum_{i=0}^{d-1} \gamma^i \left(r_i - (\tau+\lambda) \log \pistar(a_i | s_i) + \lambda \log {\ppi(a_i | s_i)} \right)\right].
\label{eq:multi2}
\end{equation}
}

\subsection{Trust-PCL}
We propose to train a parameterized policy $\pitheta$ 
and value estimate $\vphi$ to satisfy the multi-step 
consistencies in~\eqref{eq:multi2}.  
Thus, we define a consistency error for a sequence of 
states, actions, and rewards 
$s_{t:t+d} \equiv (s_t, a_t, r_t, \dots, s_{t+d-1}, a_{t+d-1}, r_{t+d-1}, s_{t+d})$ 
sampled from the environment as
\begin{equation}
\begin{aligned}
C(s_{t:t+d}, \theta, \phi) ~=~& -\vphi(s_t) + \gamma^{d} \vphi(s_{t+d})~+ \\
& \sum_{i=0}^{d-1} \gamma^i \left(r_{t+i} - (\tau +\lambda) \log\pitheta(a_{t+i} | s_{t+i}) + \lambda \log{\ppitheta(a_{t+i} | s_{t+i})}\right)~.
\end{aligned}
\label{eq:consistency-error}
\end{equation}
 
We aim to minimize the squared consistency error on every 
sub-trajectory of length $d$.  
That is, the loss for a given batch of episodes (or sub-episodes) 
$S = \{s^{(k)}_{0:T_k}\}_{k=1}^B$ is
\begin{equation}
\calL(S, \theta, \phi) = \sum_{k=1}^B \sum_{t=0}^{T_k - 1} C(s^{(k)}_{t:t+d}, \theta, \phi)^2.
\end{equation}
We perform gradient descent on $\theta$ and $\phi$ to minimize this loss.
In practice, we have found that it is
beneficial to learn the parameter $\phi$ at least as fast as $\theta$, and
accordingly, given a mini-batch of episodes we perform a single gradient
update on $\theta$ and possibly multiple gradient updates on $\phi$ (see Appendix for details).
 
In principle, the mini-batch $S$ may be taken from either on-policy or
off-policy trajectories. In our implementation, we utilized a replay
buffer prioritized by recency.  As episodes (or sub-episodes) are sampled from the
environment they are placed in a replay buffer and a priority
$p(s_{0:T})$ is given to a trajectory $s_{0:T}$ equivalent to the
current training step.  Then, to sample a batch for training, $B$
episodes are sampled from the replay buffer proportional to
exponentiated priority $\exp\{\beta p(s_{0:T})\}$ for some hyperparameter 
$\beta \ge 0$.
 
For the prior policy $\ppitheta$, we use a lagged geometric mean of
the parameters.  At each training step, we update $\tilde{\theta}
\leftarrow \alpha\tilde{\theta} + (1 - \alpha) \theta$.  Thus on
average our training scheme attempts to maximize entropy regularized
expected reward while penalizing divergence from a policy roughly $1 /
(1 - \alpha)$ training steps in the past. 

\subsection{Automatic Tuning of The Lagrange Multiplier $\lambda$}
\label{sec:lambda}

The use of a relative entropy regularizer as a penalty 
rather than a constraint introduces several difficulties.  
The hyperparameter $\lambda$ must necessarily adapt to the 
distribution of rewards.  
Thus, $\lambda$ must be tuned not only to each environment 
but also during training on a single environment, 
since the observed reward distribution changes as the agent's 
behavior policy improves.  
Using a constraint form of the regularizer is more desirable, 
and others have advocated its use in practice~\citep{trpo} 
specifically to robustly allow larger updates during training.
 
To this end, we propose to redirect the hyperparameter tuning 
from $\lambda$ to $\epsilon$.  
Specifically, we present a method which, given a desired hard 
constraint on the relative entropy defined by $\epsilon$, 
approximates the equivalent penalty coefficient $\lambda(\epsilon)$.
This is a key novelty of our work and is distinct
from previous attempts at automatically tuning 
a regularizing coefficient, which iteratively increase
and decrease the coefficient based on observed training 
behavior~\citep{ppo, heess2017emergence}.
 
We restrict our analysis to the undiscounted setting $\gamma=1$
with entropy regularizer $\tau=0$.  
Additionally, we assume deterministic, finite-horizon environment dynamics.  
An additional assumption we make is that the expected KL-divergence
over states is well-approximated by the KL-divergence starting from
the unique initial state $s_0$.
Although in our experiments these restrictive assumptions are not met, 
we still found our method to perform well for adapting 
$\lambda$ during training.  
 
In this setting the optimal policy of~\eqref{eq:objrent}
is proportional to exponentiated scaled reward.  
Specifically, for a full episode $s_{0:T} = (s_0, a_0, r_0,\dots, s_{T-1}, a_{T-1}, r_{T-1}, s_T)$, 
we have
\begin{equation}
\pistar(s_{0:T}) \propto \ppi(s_{0:T}) \exp\left\{ \frac{R(s_{0:T})}{\lambda}\right\},
\end{equation}
where $\pi(s_{0:T}) = \prod_{i=0}^{T-1} \pi(a_i | s_i)$ and 
$R(s_{0:T}) = \sum_{i=0}^{T-1} r_i$.  
The normalization factor of $\pistar$ is
\begin{equation}
Z = \expected_{s_{0:T}\sim \ppi} \left[ \exp\left\{ \frac{R(s_{0:T})}{\lambda}\right\} \right].
\end{equation}
We would like to approximate the trajectory-wide KL-divergence
between $\pistar$ and $\ppi$.
We may express the KL-divergence analytically:
\begin{align}
KL(\pistar || \ppi) &= \expected_{s_{0:T}\sim\pistar} \left[\log\left(\frac{\pistar(s_{0:T})}{\ppi(s_{0:T})}\right)\right] \\
 & =  \expected_{s_{0:T}\sim\pistar} \left[\frac{R(s_{0:T})}{\lambda} - \log Z\right] \\
& =  -\log Z + \expected_{s_{0:T}\sim\ppi} \left[\frac{R(s_{0:T})}{\lambda} \cdot \frac{\pistar(s_{0:T})}{\ppi(s_{0:T})} \right] \\
& =  -\log Z + \expected_{s_{0:T}\sim\ppi} \left[\frac{R(s_{0:T})}{\lambda} \exp\{ R(s_{0:T}) / \lambda - \log Z\} \right].
\end{align}
Since all expectations are with respect to $\ppi$,
this quantity is tractable to approximate 
given episodes sampled from $\ppi$ 
 
Therefore, in Trust-PCL, given a set of episodes sampled from 
the prior policy $\ppitheta$ and a desired 
maximum divergence $\epsilon$, 
we can perform a simple line search to find a suitable 
$\lambda(\epsilon)$ which yields 
$KL(\pistar || \ppitheta)$ as close as possible 
to $\epsilon$.
 
The preceding analysis provided a method to determine 
$\lambda(\epsilon)$ given a desired maximum divergence 
$\epsilon$.  
However, there is still a question of whether 
$\epsilon$ should change during training.  
Indeed, as episodes may possibly increase in length, 
$KL(\pistar || \ppi)$ naturally increases when 
compared to the average per-state 
$KL(\pistar(-|s) || \ppi(-|s))$, 
and vice versa for decreasing length.  
Thus, in practice, given an $\epsilon$ and a set of 
sampled episodes $S = \{s^{(k)}_{0:T_k}\}_{k=1}^N$, we 
approximate the best $\lambda$ which yields a maximum 
divergence of $\frac{\epsilon}{N} \sum_{k=1}^N T_k$.  
This makes it so that $\epsilon$ corresponds more 
to a constraint on the length-averaged KL-divergence.
 
To avoid incurring a prohibitively large number of interactions with the 
environment for each parameter update,
in practice 
we use the last 100 episodes as the set of sampled episodes $S$.  
While this is not exactly the same as sampling episodes from $\ppitheta$, 
it is not too far off since $\ppitheta$ is a lagged 
version of the online policy $\pitheta$.  
Moreover, we observed this protocol to work well in practice.  
A more sophisticated and accurate protocol may be derived 
by weighting the episodes according to the importance 
weights corresponding to their true sampling distribution.

\section{Experiments}

We evaluate Trust-PCL against TRPO
on a number of benchmark tasks.
We choose TRPO as a baseline since it is a 
standard algorithm known to achieve 
state-of-the-art
performance on the continuous control tasks
we consider (see \eg\ leaderboard results on
the OpenAI Gym website~\citep{gym}).
We find that Trust-PCL can match or improve upon
TRPO's performance in terms of both average
reward and sample efficiency.

\subsection{Setup}
We chose a number of control tasks available
from OpenAI Gym~\citep{gym}.  The first task,
Acrobot, is a discrete-control task, while
the remaining tasks (HalfCheetah, Swimmer,
Hopper, Walker2d, and Ant) are well-known
continuous-control tasks utilizing the 
MuJoCo environment~\citep{mujoco}.  

For TRPO we trained using batches of $Q=25,000$
steps ($12,500$ for Acrobot), which is the approximate 
batch size used by other implementations~\citep{rllab,modularrl}.
Thus, at each training iteration, TRPO samples
$25,000$ steps using the policy $\ppitheta$ and then
takes a single step within a KL-ball to yield a
new $\pitheta$.

Trust-PCL is off-policy, so to evaluate its performance
we alternate between collecting experience and training
on batches of experience sampled from the replay buffer.
Specifically, we alternate between collecting $P=10$ steps 
from the environment and performing a single gradient
step based on a batch of size $Q=64$ sub-episodes of length $P$
from the replay buffer,
with a recency weight of $\beta=0.001$ on the sampling distribution
of the replay buffer.
To maintain stability we use $\alpha=0.99$ and
we modified
the loss from squared loss to Huber loss on the consistency error.
Since our policy is parameterized by a unimodal Gaussian, it is impossible
for it to satisfy all path consistencies, and so we found this
crucial for stability.

For each of the variants and for each
environment, we performed a hyperparameter search to find the best
hyperparameters.  The plots presented here show the reward achieved
during training on the best hyperparameters averaged over the best
$4$ seeds of $5$ randomly seeded training runs. Note that this
reward is based on greedy actions (rather than random sampling).

Experiments were performed using Tensorflow~\citep{tensorflow}.
Although each training step of Trust-PCL (a simple gradient step)
is considerably faster than TRPO, we found that this does not have
an overall effect on the run time of our implementation, 
due to a combination of the fact
that each environment step is used in multiple training steps of Trust-PCL
and that a majority of the run time is spent interacting with the environment.
A detailed description of our implementation and hyperparameter search
is available in the Appendix.

\subsection{Results}

\begin{figure}[h]
\begin{center}
  \begin{tabular}{@{}c@{}c@{}c@{}}
    \tiny Acrobot & \tiny HalfCheetah & \tiny Swimmer \\
    \includegraphics[width=0.27\columnwidth]{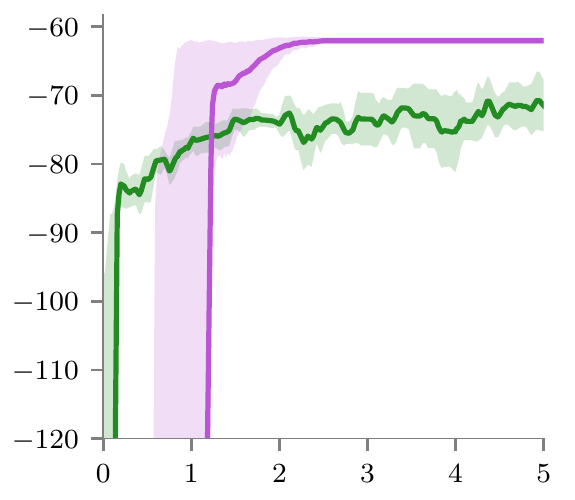} &
    \includegraphics[width=0.27\columnwidth]{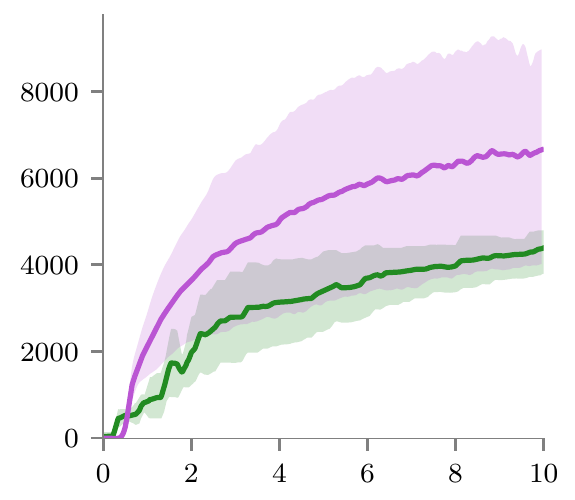} &
    \includegraphics[width=0.27\columnwidth]{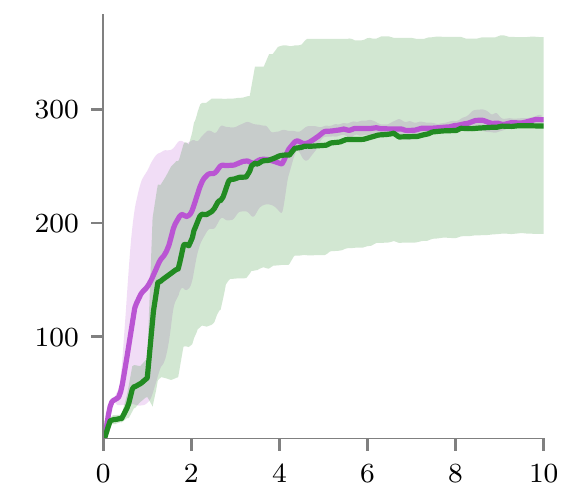} \\
    \tiny Hopper & \tiny Walker2d & \tiny Ant \\
    \includegraphics[width=0.27\columnwidth]{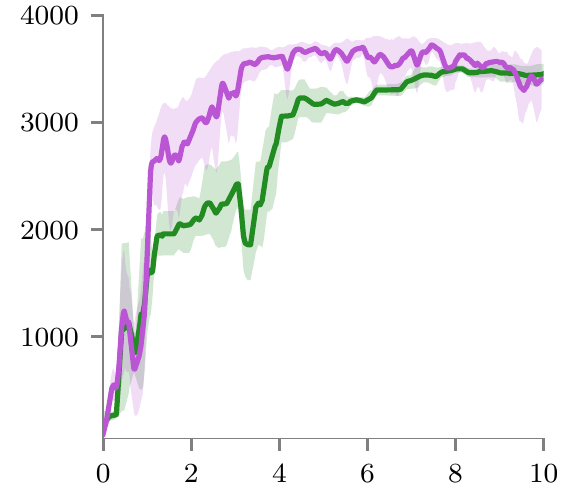} &
    \includegraphics[width=0.27\columnwidth]{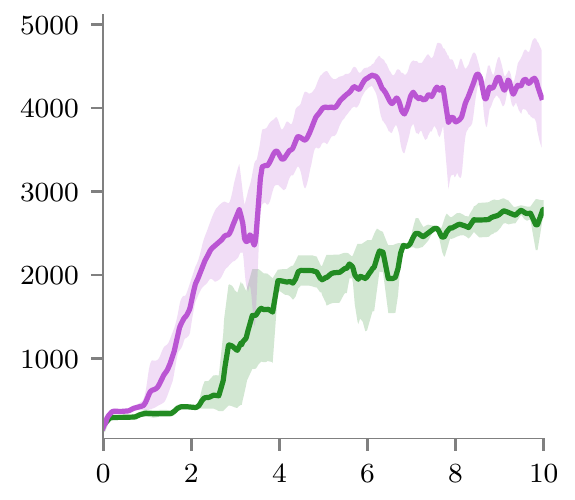} &
    \includegraphics[width=0.27\columnwidth]{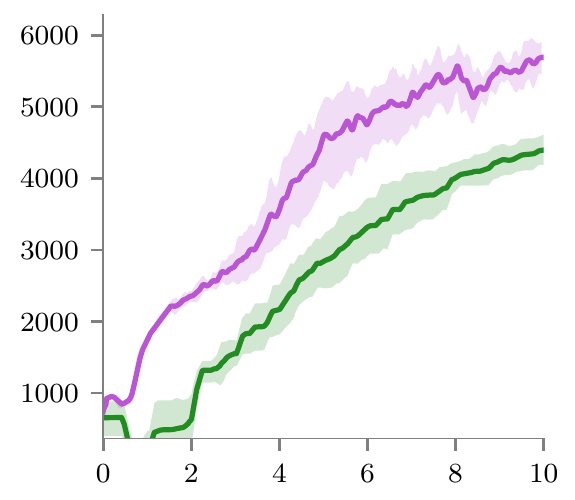} \\
    \multicolumn{3}{c}{\includegraphics[width=0.5\columnwidth]{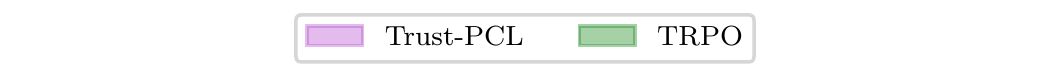}}
  \end{tabular}
\end{center}
\caption{
The results of Trust-PCL
 against a TRPO baseline.
Each plot shows average greedy reward with single standard deviation
error intervals capped at the min and max across $4$ best of $5$ 
randomly seeded training runs after choosing best hyperparameters.
The x-axis shows millions of environment steps.
We observe that Trust-PCL is consistently able to match
and, in many cases, beat TRPO's performance both in terms of 
reward and sample efficiency.
}
\label{fig:results}
\end{figure}

\begin{figure}[h]
\begin{center}
  \begin{tabular}{cc}
    \tiny Hopper & \tiny Walker2d \\
    \includegraphics[width=0.27\columnwidth]{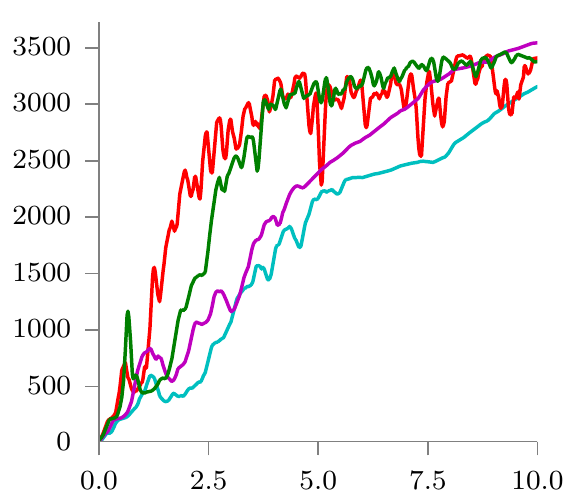} &
    \includegraphics[width=0.27\columnwidth]{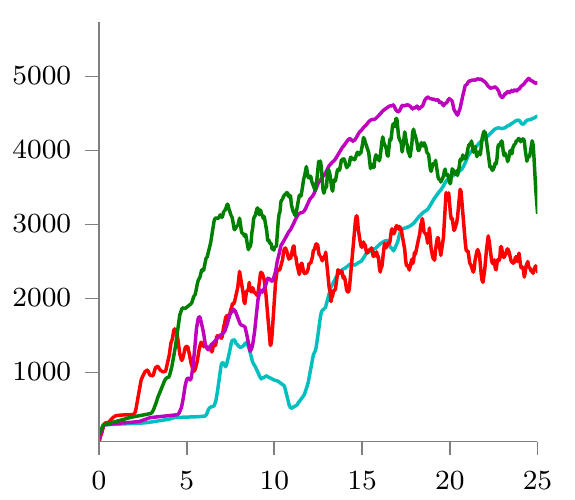} \\
    \multicolumn{2}{c}{\includegraphics[width=0.5\columnwidth]{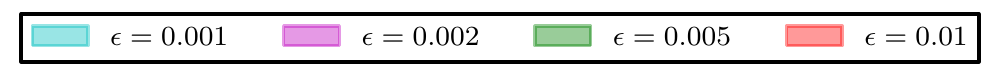}}
  \end{tabular}
\end{center}
\caption{
The results of Trust-PCL across several values of $\epsilon$,
defining the size of the trust region.
Each plot shows average greedy reward across $4$ best of $5$ 
randomly seeded training runs after choosing best hyperparameters.
The x-axis shows millions of environment steps.
We observe that instability increases with $\epsilon$,
thus concluding that the use of trust region is crucial.
}
\label{fig:epsilon}
\end{figure}

\begin{figure}[h]
\begin{center}
  \begin{tabular}{cc}
    \tiny Hopper & \tiny Walker2d \\
    \includegraphics[width=0.27\columnwidth]{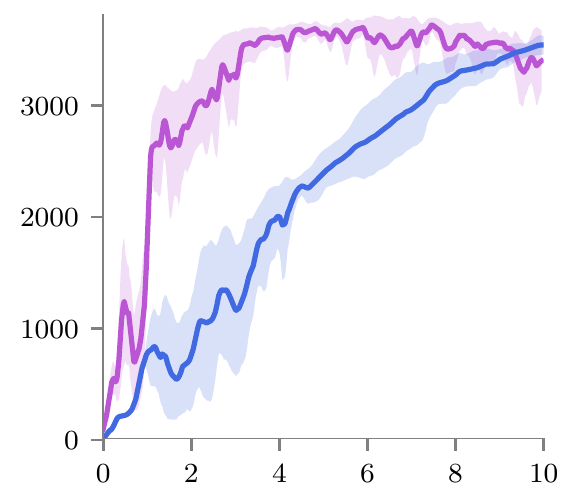} &
    \includegraphics[width=0.27\columnwidth]{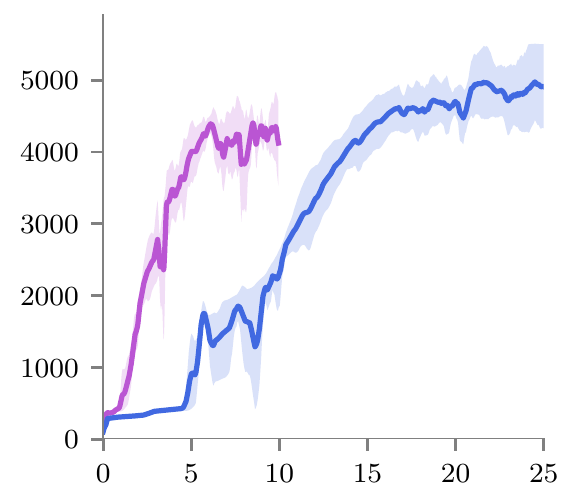} \\
    \multicolumn{2}{c}{\includegraphics[width=0.5\columnwidth]{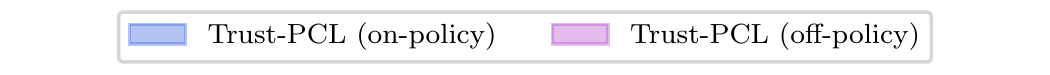}}
  \end{tabular}
\end{center}
\caption{
The results of Trust-PCL varying the degree of on/off-policy.
We see that Trust-PCL (on-policy) has a behavior similar to TRPO,
achieving good final reward but requiring an exorbitant number
of experience collection.  When collecting less experience
per training step in Trust-PCL (off-policy), we are able
to improve sample efficiency while still achieving
a competitive final reward.
}
\label{fig:onoff-policy}
\end{figure}

\begin{table*}[t]
	\centering
    {\small
    \begin{tabular}{c | c | c | c | c | c}
        \hline
        {\bf Domain} & {\bf TRPO-GAE} & {\bf TRPO (rllab)} &
        {\bf TRPO (ours)} & {\bf Trust-PCL} & {\bf IPG} \\
        \hline
        HalfCheetah & 4871.36 & 2889 & 4343.6 & {\bf 7057.1 } & 4767 \\
        Swimmer & 137.25 & -- & 288.1 & {\bf 297.0 } & -- \\
        Hopper & 3765.78 & -- & 3516.7 &  {\bf 3804.9 } & -- \\
        Walker2d & {\bf 6028.73 } & 1487 & 2838.4 & 5027.2 & 3047 \\
        Ant & 2918.25 & 1520 & 4347.5 & {\bf 6104.2} & 4415 \\
    \end{tabular}
    }
\caption{
Results for best average reward in the first 10M steps of training
for our implementations (TRPO (ours) and Trust-PCL) and external implementations.
TRPO-GAE are results of~\citet{modularrl} available on the OpenAI Gym website.
TRPO (rllab) and IPG are taken from~\citet{gu2017interpolated}.
These results are each on different setups with different hyperparameter
searches and in some cases different evaluation protocols
(\eg TRPO (rllab) and IPG were run with a simple linear value network instead of the
two-hidden layer network we use).  
Thus, it is not possible to make any definitive claims based on this data.
However, we do conclude that our results are overall competitive with state-of-the-art
external implementations.
}
\label{tab:average_reward}
\end{table*}

\comment{
\begin{table*}[t]
	\centering
    {\small
    \begin{tabular}{c | c | c}
        \hline
        {\bf Algorithm} & {\bf Ant} & {\bf Half-Cheetah}\\
        \hline
        IPG-$\nu$=0.2-$\beta$ & 4299.62 & 3433.98 \\
        IPG-$\nu$=0.2-$\beta$-cv & 2783.11 & 3954.93 \\
        IPG-$\nu$=0.2-$\pi$ & 3944.77 & 3331.81 \\
        IPG-$\nu$=0.2-$\pi$-cv & 3893.32 & 4161.79 \\
        TRPO-05000- & 1494.97 & 2785.28 \\ 
    \end{tabular}
    }
   	\caption{Shane's numbers, final average reward after 10,000 episodes over 3 seeds}
    \label{tab:interpolated_pg}
\end{table*}
}

We present the reward over training of Trust-PCL and TRPO
in~\figref{fig:results}.  
We find that Trust-PCL can match or beat the performance of TRPO
across all environments in terms of both final reward and sample
efficiency.
These results are especially significant on the harder tasks
(Walker2d and Ant).
We additionally present our results compared to other published results 
in~\tabref{tab:average_reward}.  We find that even when comparing across
different implementations, Trust-PCL can match or beat the state-of-the-art.

\subsubsection{Hyperparameter Analysis}
The most important hyperparameter in our method is $\epsilon$, which 
determines the size of the trust region and thus has a critical
role in the stability of the algorithm.
To showcase this effect, 
we present the reward during training for several different values 
of $\epsilon$ in~\figref{fig:epsilon}.  As $\epsilon$ increases,
instability increases as well, eventually having an adverse effect
on the agent's ability to achieve optimal reward.  
Note that standard PCL~\citep{pcl} corresponds to $\epsilon\to\infty$
(that is, $\lambda = 0$).  
Therefore, standard PCL would fail in these environments,
and the use of trust region is crucial.

The main advantage of Trust-PCL over existing trust region
methods for continuous control is its ability to learn
in an off-policy manner.  
The degree to which Trust-PCL is off-policy is determined by a combination
of the hyparparameters $\alpha$, $\beta$, and $P$.
To evaluate the importance of training
off-policy, we evaluate Trust-PCL with a hyperparameter setting
that is more on-policy.  We set $\alpha=0.95$, $\beta=0.1$, and $P=1,000$.
In this setting, we also use large batches of $Q=25$ episodes of length $P$
(a total of $25,000$ environment steps per batch).
\figref{fig:onoff-policy} shows the results of Trust-PCL 
with our original parameters and this new setting.
We note a
dramatic advantage in sample efficiency when using 
off-policy training.
Although Trust-PCL (on-policy) can achieve state-of-the-art
reward performance, it requires an exorbitant amount of experience.
On the other hand, Trust-PCL (off-policy) can be competitive in
terms of reward while providing a significant improvement
in sample efficiency.

One last hyperparameter is $\tau$, determining the 
degree of exploration.  Anecdotally, we found $\tau$
to not be of high importance for the tasks we evaluated.
Indeed many of our best results use $\tau=0$.  Including $\tau>0$
had a marginal effect, at best.  
The reason for this is likely due to the tasks themselves.
Indeed, other works which focus on exploration in continuous control
have found the need to propose 
exploration-advanageous variants of these standard 
benchmarks~\citep{haarnoja2017reinforcement, vime}.

\section{Conclusion}

We have presented Trust-PCL, an off-policy algorithm employing
a relative-entropy penalty to impose a trust region on
a maximum reward objective.  We found that Trust-PCL
can perform well on a set of standard control tasks,
improving upon TRPO both in terms of average reward
and sample efficiency.  
Our best results on Trust-PCL are able to maintain
the stability and solution quality of TRPO
while approaching the sample-efficiency of
value-based methods (see \eg~\citet{lmetz}).
This gives hope that the goal of achieving
both stability and sample-efficiency without
trading-off one for the other is attainable
in a single unifying RL algorithm.

\section{Acknowledgment}

We thank Matthew Johnson, Luke Metz, Shane Gu, and the Google Brain
team for insightful comments and discussions.

\bibliography{main.bib}
\bibliographystyle{iclr2018_conference}

\newpage
\appendix

\section{Implementation Benefits of Trust-PCL}

We have already highlighted the ability of Trust-PCL 
to use off-policy data to stably train both a 
parameterized policy and value estimate, 
which sets it apart from previous methods.  
We have also noted the ease with which exploration can 
be incorporated through the entropy regularizer.  
We elaborate on several additional benefits of Trust-PCL.
 
Compared to TRPO, Trust-PCL is much easier to implement.  
Standard TRPO implementations perform second-order 
gradient calculations on the KL-divergence to construct 
a Fisher information matrix (more specifically a vector 
product with the inverse Fisher information matrix).  
This yields a vector direction for which a line search is 
subsequently employed to find the optimal step.
Compare this to Trust-PCL which employs simple gradient descent.  
This makes implementation much more straightforward 
and easily realizable within standard deep learning frameworks.
 
Even if one replaces the constraint on the average 
KL-divergence of TRPO with a simple regularization penalty 
(as in proximal policy gradient methods \citep{ppo, acer}), 
optimizing the resulting objective requires computing 
the gradient of the KL-divergence.  
In Trust-PCL, there is no such necessity.  
The per-state KL-divergence need not have an 
analytically computable gradient.  
In fact, the KL-divergence need not have a closed form at all.  
The only requirement of Trust-PCL is that the 
log-density be analytically computable. 
This opens up the possible policy parameterizations 
to a much wider class of functions.  
While continuous control has traditionally used 
policies parameterized by unimodal Gaussians, 
with Trust-PCL the policy can be replaced with 
something much more expressive---for example, 
mixtures of Gaussians or auto-regressive policies
as in~\citet{lmetz}.
 
We have yet to fully explore these additional benefits in this work, 
but we hope that future investigations can exploit the flexibility and ease 
of implementation of Trust-PCL to further the progress of RL 
in continuous control environments.

\section{Experimental Setup}

We describe in detail the experimental setup regarding
implementation and hyperparameter search.

\subsection{Environments}
In Acrobot, episodes were cut-off at step $500$.
For the remaining environments, episodes were cut-off at step $1,000$.

Acrobot, HalfCheetah, and Swimmer are all non-terminating environments.
Thus, for these environments, each episode had equal length and each 
batch contained the same number of episodes.
Hopper, Walker2d, and Ant are environments that can terminate
the agent.  Thus, for these environments, the batch size
throughout training
remained constant in terms of steps but not in terms
of episodes.

There exists an additional common MuJoCo task called Humanoid.
We found that neither our implementation of TRPO nor Trust-PCL could make more than negligible
headway on this task, and so omit it from the results.
We are aware that TRPO with the addition of GAE and enough fine-tuning
can be made to achieve good results on Humanoid~\citep{trpo_gae}.
We decided to not pursue a GAE implementation to keep a fair comparison
between variants.  Trust-PCL can also be made to incorporate an
analogue to GAE (by maintaining consistencies at varying time scales),
but we leave this to future work.

\subsection{Implementation Details}
We use fully-connected feed-forward neural networks
to represent both policy and value.  

The policy $\pitheta$ is
represented by a neural network with two hidden layers
of dimension $64$ with $\tanh$ activations.  At time step $t$,
the network
is given the observation $s_t$.
It produces a vector $\mu_t$, which is combined with a learnable
(but $t$-agnostic) parameter $\xi$ to parametrize a unimodal
Gaussian with mean $\mu_t$ and standard deviation $\exp(\xi)$.
The next action $a_t$ is sampled randomly from this Gaussian.

The value network $\vphi$ is represented by a neural
network with two
hidden layers of dimension $64$ with $\tanh$ activations.
At time step $t$
the network is given the observation $s_t$ and the component-wise
squared observation $s_t\odot s_t$.  It produces a single scalar value.

\subsubsection{TRPO Learning}
At each training iteration, both the policy and value parameters are updated.
The policy is trained by performing a trust region step
according to the procedure described in~\citet{trpo}.

The value parameters at each step are solved
using an LBFGS optimizer.  To avoid instability, the value parameters
are solved to fit a mixture of the empirical values and the expected values.
That is, we determine $\phi$ to minimize
$\sum_{s\in\text{batch}} (\vphi(s) - \kappa V_{\tilde{\phi}}(s) - (1-\kappa)\hat{V}_{\tilde{\phi}}(s))^2$,
where again $\tilde{\phi}$ is the previous value parameterization.
We use $\kappa=0.9$.  This method for training $\phi$ is according to that used in~\citet{modularrl}.

\subsubsection{Trust-PCL Learning}
At each training iteration, both the policy and value parameters are updated.
The specific updates are slightly different between \tpcla and \tpclb.

For \tpcla, the policy is trained by taking a single gradient step using
the Adam optimizer~\citep{adam} with learning rate $0.001$.
The value network update is inspired by that used in TRPO
we perform 5 gradients steps with learning rate $0.001$, 
calculated with regards to a mix between the empirical values
and the expected values according to the previous $\tilde{\phi}$.  We use $\kappa=0.95$.

For \tpclb, both the policy and value parameters are updated in a single step
using the Adam optimizer with learning rate $0.0001$.  For this variant,
we also utilize a target value network (lagged at the same rate as the
target policy network) to replace the value estimate at the final
state for each path.  We do not mix between empirical and expected values.

\subsection{Hyperparameter Search}
We found the most crucial hyperparameters for effective learning
in both TRPO and Trust-PCL
to be $\epsilon$ (the constraint defining the size of the trust region)
and $d$ (the rollout determining how to evaluate the empirical value of a state).
For TRPO we performed a grid search over 
$\epsilon\in\{0.01, 0.02, 0.05, 0.1\}, d\in\{10, 50\}$.
For Trust-PCL we performed a grid search over 
$\epsilon\in\{0.001, 0.002, 0.005, 0.01\}, d\in\{10, 50\}$.
For Trust-PCL we also experimented with the value of $\tau$, either keeping
it at a constant 0 (thus, no exploration) or decaying it from $0.1$ to $0.0$
by a smoothed exponential rate of $0.1$ every 2,500 training iterations.

We fix the discount to $\gamma=0.995$ for all environments.

\section{Pseudocode}
A simplified pseudocode for Trust-PCL is presented in
Algorithm~\ref{alg:tpcl}.
\begin{algorithm}[!htb]
\caption{Trust-PCL}
\label{alg:tpcl}

\begin{algorithmic}
\STATE {\bfseries Input:} 
Environment $ENV$, trust region constraint $\epsilon$,
learning rates $\eta_\pi,\eta_v$, discount factor $\gamma$,
rollout $d$, batch size $Q$, collect steps per train step $P$, 
number of training steps $N$, replay buffer $RB$
with exponential lag $\beta$, lag on prior policy $\alpha$.

\STATE
\FUNCTION{Gradients($\{s^{(k)}_{t:t+P}\}_{k=1}^B$)}
\STATE \emph{// $C$ is the consistency error defined in Equation~\ref{eq:consistency-error}.}
\STATE Compute 
$\Delta\theta = \sum_{k=1}^B \sum_{p=0}^{P-1} C(s^{(k)}_{t+p:t+p+d}, \theta, \phi) 
\nabla_\theta C(s^{(k)}_{t+p:t+p+d}, \theta, \phi)$.
\STATE Compute 
$\Delta\phi = \sum_{k=1}^B \sum_{p=0}^{P-1} C(s^{(k)}_{t+p:t+p+d}, \theta, \phi) 
\nabla_\phi C(s^{(k)}_{t+p:t+p+d}, \theta, \phi)$.
\STATE \emph{Return} $\Delta\theta, \Delta\phi$
\ENDFUNCTION
\STATE
\STATE Initialize $\theta,\phi,\lambda$, set $\tilde{\theta} = \theta$.
\STATE Initialize empty replay buffer $RB(\beta)$.
\FOR{$i=0$ {\bfseries to} $N-1$}
\STATE \emph{// Collect}
\STATE Sample $P$ steps $s_{t:t+P}\sim\pitheta$ on $ENV$.
\STATE Insert $s_{t:t+P}$ to $RB$.

\STATE
\STATE \emph{// Train}
\STATE Sample batch $\{s^{(k)}_{t:t+P}\}_{k=1}^B$ from $RB$ to contain a total of $Q$ transitions ($B\approx Q / P$).
\STATE $\Delta\theta, \Delta\phi = \text{Gradients}(\{s^{(k)}_{t:t+P}\}_{k=1}^B)$.
\STATE Update $\theta \leftarrow \theta - \eta_\pi\Delta\theta$.
\STATE Update $\phi \leftarrow \phi - \eta_v\Delta\phi$.

\STATE
\STATE \emph{// Update auxiliary variables}
\STATE Update $\tilde{\theta} = \alpha\tilde{\theta} + (1 - \alpha) \theta$.
\STATE Update $\lambda$ in terms of $\epsilon$ according to Section~\ref{sec:lambda}.

\ENDFOR

\end{algorithmic}
\end{algorithm}

\end{document}